**Factors other than climate change are currently more important in predicting how well fruit farms are doing financially.**

**Fabian Obster[1,2], Heidi Bohle [3] and Paul M. Pechan[3*]**

[1] Department of Business Administration, Bundeswehr University Munich, Munich, Germany
[2] Department of Statistics, Ludwig Maximilian University Munich, Germany
[3] Institute of Communication and Media Research, Ludwig Maximilian University Munich, Germany
*Corresponding author

## Key words

## Abstract

Machine learning and statistical modeling methods were used to analyze the impact of climate change on financial wellbeing of fruit farmers in Tunisia and Chile. The analysis was based on face to face interviews with 801 farmers. Three research questions were investigated. First, whether climate change impacts had an effect on how well the farm was doing financially. Second, if climate change was not influential, what factors were important for predicting farm's financial wellbeing. And third, ascertain whether observed effects on the financial wellbeing of the farm were a result of interactions between predictor variables. This is the first report directly comparing climate change with other factors potentially impacting financial wellbeing of (fruit) farms. Certain climate change factors, namely increases in temperature and reductions in precipitation, can regionally impact self-perceived financial wellbeing of fruit farmers. Specifically, increases in temperature and reduction in precipitation can have a measurable negative impact on the financial wellbeing of farms in Chile. This effect is less pronounced in Tunisia. Climate impact differences were observed within Chile but not in Tunisia. However, climate change is only of minor importance for predicting farm financial wellbeing, especially for farms already doing financially well. Factors that are more important, mainly in Tunisia, included trust in information sources and prior farm ownership. Other important factors include farm size, water management systems used and diversity of fruit crops grown. Moreover, some of the important factors identified differed between farms doing and not doing well financially. Interactions between factors may improve or worsen farm financial wellbeing. The results illustrate the usefulness of combining supervised machine learning with classical statistical tools for data analysis in the agronomical sector.

## Introduction

### Climate change impact on fruit tree yields and farm economic wellbeing

Climate change can impact crops, with yields of many important crops projected to decline in the future (1,2). For example, increases in temperature can reduce yields of major crops worldwide (3). Such climatic impacts can and will have a detrimental effect on the availability and nutritional value of food (4). In this paper we focus on the effects of climate change on crops that have an important nutritional and monetary value in Chile and Tunisia: cherry and peach fruit tree (5,6). Reproductive organs of both fruit crops can be damaged by climate change leading to reduction in quantity and quality of harvestable fruit (7,8,9). Thus increases in winter temperatures can affect fruit tree chill requirements resulting in changes of bud, flower and fruit set (10,11,12,13,14). Similarly, increasing temperatures during the fruit set and development can lead to changes in fruit growth and maturation (9,15,16). In combination with reduced water availability, high temperatures can affect both fruit yield and quality (7,17,18). These effects can vary between fruit tree cultivars and species.  Extreme events (hail, wind, frost) have also been observed to impact the physical environment and cause fruit crop damage (7,19,20,21).  These climate events are region specific, affecting food production the crops to different extents.

Although climate change impacts on fruit crop quality and yield can be estimated, making predictions/projections about climate change impacts on farm incomes (revenues) is much more difficult and results in a high degree of outcome uncertainty (22). Yet the economic estimates are crucial, as it is primarily the farmers, using their own financial resources, who need to take the necessary adaptive measures to reduce the impact of climate change (or any other potential harm) on their crops. Most economic climate change impact models are based on estimating impacts of climate change on crop yields (based on climate and crop simulation models) and translating this information into likely farm financial performance. However, such analysis is based on a number of assumptions and seldom takes a combination of adaptive measures, socio-economic and other factors, such as regional differences, into consideration (23).  One of the most often used economic models measuring impacts of climate change on agriculture is the Ricardian approach that focuses on the land value and agricultural revenue (24,25). Cross-sectional and panel regression analysis are the analytical tools of choice (26). Whatever the approach and type of analysis performed, omission of variables that may directly or indirectly affect crop/farm incomes/revenue makes climate change financial impact assessments highly uncertain. The ability to estimate variable importance for the economic well being of a farm while presenting overall predictability values for the analysis would go a long way to reduce this uncertainty.

**Application of machine learning in agriculture**

Predictive analysis generates predictions about the future based on available data. In classical statistics, regression analysis can have predictive powers. There are however situations where regression analysis is not sufficient to handle the generated datasets or the specific questions to be answered or where the assumption of the existence of a linear function between independent and dependent variables doesn't hold. This is especially the case when complex variable interactions are present in the dataset. Here, artificial intelligence (AI), and machine learning in particular, become useful tools that complement traditional statistical analysis (27,28,29). Apart from its complementary role, machine learning, because of its ability to analyse large datasets and many variables simultaneously, can help to reduce the chance that important variables are left out of the data analysis process.

It comes thus as no surprise that machine learning can offer new possibilities to analyse big data in agriculture (30,31,32,33,34,35,36,37). Indeed, the list of possible applications of machine learning to agriculture is potentially vast, especially when considered in combination with other research domains, such as climate change (38,39,40,41). However, tackling agricultural problems is complex. For example, whether a new crop variety actually provides better yield and farm income under certain climatic conditions is potentially dependent not only on its genetic traits but also on many other factors, such as those related to biophysical and farm management issues (42). This means that complex and deep interactions could exist in the datasets. Such data can become quickly difficult to properly analyze using classical statistical approaches. This is where the power of machine learning can be explored to its full potential (36,43). Not only can biophysical variables such as microclimate effects, soil structure and quality be included in the analysis, but also socio-economic variables, such as local land use, urban-farm water accessibility, water-related policies, farm size, demographic data and access to markets can also be included, allowing analysis at every step of the agricultural value chain (32, 44, 45). The resulting datasets, just for one farm, could encompass millions of data point combinations. Importantly, analysis of such data can provide answers as to whether adaptive measures were effective in maintaining or increasing crop yield and farm incomes under certain climatic conditions, provide information on the relative importance of the intervention to the desired outcome and help with future (yield) predictions (38, 40, 46, 47, 48). Unfortunately, the use of the machine learning approach on its own can limit data interpretability. This is the case for complex machine learning algorithms, such as support vector machines, deep neural networks, and random forest or boosted trees. Even though there are post hoc interpretability methods to approximate the functioning of such black box models, there is no straightforward way of understanding and interpreting the exact processes leading to the outcome. This is potentially a major drawback for research questions that aim to deepen the understanding of the processes or factors associated with the desired outcome. To overcome this problem, we introduce herein a (hybrid) method that combines analysis of datasets based on generalized linear models combined with strategies from machine learning, such as cross-validation and boosting and group-variable selection. The output of this approach preserves interpretability, respects the group structure of the data and is still competitive with state-of-the-art machine learning algorithms. Details of this strategy are presented in the data analysis section of this paper.

**Research approach and objectives**

The aim of this paper is to evaluate likely climate change effects on fruit farm financial wellbeing. The approach taken was to rely on farmer self-reporting about the past experiences with climate change and whether these experiences had any bearing on the financial wellbeing of the farm. The information was collected through face to face interviews. Because it was the farmers who provided the information for subsequent data analysis, we are in effect reporting herein on farmer´s perceived financial wellbeing.

We have utilized a combination of classical statistical analysis and machine learning to take into account the high-dimensionality of the data sets, including many grouped independent variables, necessary for this type of analysis, the latter includes a novel approach that allows analysis of single and groups of independent variables to ascertain their relative importance and predictive power for the outcome variable.

In this paper, we answer three research questions. First, whether climate change had an effect on how well the farm was doing financially. Second, in case climate change was not important for the farm financial wellbeing, what factors may influence this outcome variable. And third, whether factor interactions affect farm financial wellbeing.

**Methods**

**Study Area**

Four contrasting geographical and climatic regions were selected for the study, two regions in Tunisia and two in Chile. In Tunisia, these were the Mornag and Reueb peach growing regions.

In Chile, these were the Rengo and Chillán cherry-growing regions. The selected regions in each country differed in the level and type of climate impact exposure.

## Data collection

The two data collection instruments used in this study were a) focus meetings and b) a face-to-face survey with cherry farmers in Chile and peach farmers in Tunisia. In order to be selected for the focus meetings and survey, farmers had to own the farm, manage and work on the farm and derive over 70% of their income from their farming activities.

## Survey

### Questionnaire preparation

The questionnaire for the survey was prepared in English and translated into Tunisian Arabic and Chilean Spanish. The translated documents were back-translated into English to check for inconsistencies. The survey was pre-tested with 12 farmers in consultation with Qualitas AgroConsultores in Chile and Elka Consulting in Tunisia. Based on their feedback, and that of our research colleagues in Tunisia and Chile, some questions were removed while others were reformulated.

### Interview methodology

Randomly selected cherry and peach farmers in the selected regions of Tunisia and Chile were pre-screened to fulfill the data collection eligibility criteria (see above). A total of 801 face-to-face interviews were subsequently conducted with farmers who fulfilled the selection criteria – 401 peach farmers in Tunisia (201 in  Mornag and 200 in Regueb regions) and 400 cherry farmers in Chile (200 in Rengo and 200 in  Chillán regions). The approximately one-hour-long interviews were carried out with farmers directly on their farms. The interviews were carried out after harvest completion in the fall of 2018 by Elka Consulting in Tunisia and in the spring 2019 by Qualitas AgroConsultores in Chile. Guidance was sought from our institute (ICMR, LMU) about the survey implementation and data use that included participation of human subjects. The farm data was collected according to data collection procedures applicable in each country. Informed consent for the data collection was provided by the survey participants. No personality identifiable data was collected, assuring full anonymity.

### Measurements

In the face-to-face interviews, farmers were asked to answer a combination of multiple-choice, open, Likert Scale and Yes / No questions related to climate change and climate impacts on their farms between the years 2009 and 2018 and to their past, present and planned adaptive measures. After compiling the data from farmer interviews, the resultant datasets were checked for errors and integrated into excel formats for further data analysis.

**Data analysis**

### Variables to be analysed

Climate change is a potential threat to farm financial wellbeing in Chile and Tunisia. Four different climate change factors were considered: temperature, precipitation, extreme weather and drought.

- Temperature. Farmers were asked whether the temperature over the past 10 years was increasing, decreasing, staying the same or became unpredictable.
- Precipitation. Farmers were asked whether precipitation over the past 10 years was increasing, decreasing, staying the same or became unpredictable.
- Extreme weather. Farmers were asked whether the occurrence of wind, hail or frost extreme weather events over the past 10 years was increasing, decreasing, staying the same or became unpredictable.
- Drought. Farmers were asked whether the occurrence of dry periods or drought over the past 10 years was increasing, decreasing, staying the same or became unpredictable.

In addition to climate change, there may be groups or individual farm-related variables that may, by themselves or in interaction with climate threat, affect farm financial wellbeing. An overview of all variables, corresponding groups and survey questions used in the analysis can be found in Extended Data Figures 1 and 2.

- **Groups of variables.** We have focused our analysis on groups of farm variables (assets) that may be important for farm financial wellbeing. These were:
    - Natural (geographical regions)
    - human (education, age, gender, knowledge)
    - social (reliance on/use of information, trust in information sources, community, science or religion)
    - biophysical/manufactured (farm size, water management systems used on the farm, diversity of crops used, adaptive measures)
    - economic (farm debt, farm performance, reliance on orchard income)
    - climate experience
    - income damage

The choice of the above variables was made on the basis of the five resource/capital sustainability model that addresses the concept of sustainable wealth creation (49,50).

- **Individual variables.** Above listed grouped variables were also assessed individually. In addition, other variables were examined that may, by themselves or in interaction with climate threat, affect farm financial wellbeing.
- **Dependent variable.** The outcome variable "financial wellbeing of a farm" consists of three categories. Doing well and very well, neither doing or not doing well ("neutral"), and not doing well or not well at all. Throughout the analysis, the financial wellbeing variable is coded as two separate variables. We refer to the first variable as "high wellbeing" comparing farmers who are doing well and very well financially with farmers who are doing neutral or not well (reference category) and the second one as "low wellbeing" differentiating between farmers who are not doing well financially with farmers who are doing neutral, well or very well (reference category). This enabled us to differentiate between the process leading to farmers not doing well and the process leading to farmers doing well, as the farmers who are neither doing or not doing well are always part of the reference category.

### Analysis strategy

**Research question one: does past experience with climate change affect the financial wellbeing of a farm.**

We used a statistical approach to determine the effect of independent variables on the farm financial wellbeing. As the two outcome variable "high wellbeing" and "low wellbeing" are binary, we used logistic regression and analysed the odds ratios as well as associated p-values and confidence intervals of adaptive measures and past experience for the outcome.

**Research question two: what factors, other than climate change, may be important for the financial wellbeing of a farm.**

This research question imposes a major challenge. There are many possible influencing variables in the dataset. Some may be relevant for the outcome variable, but others may not. Variables not related to the outcome variable create unnecessary background "noise" because generalized linear models tend to over-adapt to the data (the so-called overfitting) in high-dimensional cases. In the extreme case, where the number of independent variables is higher than the number of observations, linear models cannot be fitted. The solution to this problem is to perform variable selection, and then include only these variables in the model. The current practice is to perform this selection based on literature and expert knowledge. In fact, there is always an implicit variable selection process based on which such data is collected. However, one may still end up with a large number of possible influencing variables. In this situation, combination of statistics and machine learning can be used to perform the variable selection. We used model-based boosting (51), but other strategies, such as the Lasso (52) can be utilized. The model-based boosting strategy is to improve a given model by only adding variables that improve the overall model the most. The process of adding variables is stopped if a further update would not result in a "better" model.

Importantly, in some instances, grouped variables may be more important for the model than individual variables. We used sparse group boosting for this purpose (53). In sparse group boosting, the model can decide between individual variables and groups of variables. New hypotheses can be generated about the association of selected variables or groups of variables and the farm's financial wellbeing. Being able to differentiate between the importance of groups and individual variables may help in designing questionnaires because if individual variables are more important than the group, only the important individual variables need to be included in the questionnaire. This may greatly shorten the questionnaire without loss of information. Conversely, variable groups may provide information about variable interactions.

**Research question three: Are observed effects on the financial wellbeing of a farm the result of moderating effects and/or more complex relationships between variables.**
We analysed (pairwise) interaction effects of all variables on the financial wellbeing of the farm. Interactions of variables were evaluated with the help of model-based boosting, allowing comparisons of their relative importance for the outcome variable. Note that if there are p variables in the dataset, then there are $0.5* p*(p-1)$ possible interactions in the dataset, leading to an even higher dimensional noise problem. However, this brute force method has the potential to identify important moderation or additive variable effects, and thus increase our understanding of the processes leading up to the outcome.

Depending on the research question being asked, the complexity of data analysis, as described above, may still not be sufficient. In such situations, non-interpretable black-box machine learning models should be used. Comparing the predictive performance of these machine learning models with the interpretable hybrid and statistical models gives an indication of the necessary analytical complexity. If the hybrid model outperforms the black-box model regarding the predictive power (i.e. delivers better AUC), then further complexities are not necessary. If the converse is true, the goal of future research should be to understand how these complexities can be explained, for example by using highly non-linear relationships or higher-order interactions.

**Models used for data evaluation**
**Statistical models**

We used generalized linear models (54) to answer whether interventions had an impact on the outcome of interest. As the outcome variables were binary, logistic regression was used to provide odds ratios, the corresponding p-values, and confidence intervals.

**Machine Learning**

We have compared different popular machine learning models to ensure that the models used for our analysis were competitive in their predictability. A list of all models used is given in Extended Data Figure 3. In contrast to the model-based boosting models and the logistic regression, these machine learning models do not allow insight into the data.

**Hybrid Statistical - Machine learning based predictive models**

We decided to use model-based boosting as means to select variables for the predictive models. The number of boosting iterations was controlled by 25-fold cross-validation using the training data. This hyper-parameter controls effect penalization (smoothness) and regularization (variable selection) (51). Variable selection was completed in under 4000 iterations. The effect sizes, in our cases the odds ratios, were shrunken to zero through ridge regularization. This makes it easier to interpret the results since only the most important variables for the outcome must be analyzed and irrelevant variables are not considered by the model. Since the influencing variables can be clustered into groups, as described in the contextual definitions, we used sparse group boosting (53) as an extension of model-based boosting. The chosen approach allows the resulting model and variables to be interpreted similarly to generalized linear models (55). A possible alternative for this approach is to use the lasso and the sparse group lasso (56).

**Model evaluation**

70 percent of the observations in the data were randomly assigned to the training dataset and the remaining 30 percent were assigned to the test data set for the final evaluation.

Model evaluation was based on the area under the receiver operator curve, as evaluated on the test data. For the binary outcome variables, two major performance metrics were evaluated at every threshold of probability. First, the rate of correctly identified farms doing well financially, and second, the rate of correctly identified farms not doing well financially yielding the receiver operator curve (ROC). The area under the ROC (AUC) takes both rates into account by considering all possible thresholds of probabilities computed by a prediction model. We also computed the Accuracy as additional metric, which is the percentage of all correctly identified/predicted farmers in the test data set by a classification model. Even though this metric does not balance the true positive and true negative rate in unbalanced data like the AUC, it is used because of its intuitive interpretation property.

All data analyses were performed using the statistical programming environment R, visualizations were created with the R package ggplot2 (57).

## Results

### Choice of predictive models for data evaluation

Before embarking to answer the three research questions, we compared different predictive models to ascertain which model has the best predictive power and should therefore be used for the data analysis (Table 1). Except for Chile and Tunisia combined low financial wellness, the random forest (rf) tended to outperform all other models for Chile and Tunisia combined as well as for Chile and Tunisia separately. The overview of ROC curves for selected models can be found in Extended Data Figure 4 Boosted decision trees (gbm) performed similarly to sparse group boosting (sqb) and model-based boosting (mb). In all cases, neural networks (nn) performed worse than sgb and mb. Generalized linear model (glm), which consisted only of experiences with climate change and its financial impact, had lower predictive properties than sgb and mb. However, when the glm was fitted with boosting (model-based boosting-mb), which included more variables related to the farm vulnerability to climate change and geographical location, the accuracy and AUC tended to improve compared to the glm only. Including interactions between all independent variables (mb-int) did not improve the predictive outcomes of model-based boosting. The results imply that only considering experiences with climate change and its financial impact as in the glm is not enough to explain both financial wellbeing variables. Thus, additional variables had to be considered. When compared to the interpretable models, accounting for deep interactions and complex relationships like the random forest could, in some cases, result in marginal improvements in accuracy and AUC predicting high wellbeing, but for predicting low wellbeing the simpler models seem to suffice. Since our investigation necessitated data interpretation, sgb was chosen for subsequent data analysis.

| Accuracy | wellbeing | Sgb | mb | Mb int | glm | rf | nn | gbm |
|----------|-----------|-----|-----|--------|-----|-----|-----|-----|
| Chile | High | 0.65 | 0.675 | | 0.642 | 0.733 | 0.575 | 0.683 |
| Chile | Low | 0.833 | 0.833 | | 0.833 | 0.842 | 0.833 | 0.817 |
| Chile & Tunisia | High | 0.71 | 0.693 | 0.685 | 0.618 | 0.734 | 0.556 | 0.705 |
| Chile & Tunisia | Low | 0.809 | 0.809 | | 0.822 | 0.817 | 0.822 | 0.793 |
| Tunisia | High | 0.595 | 0.579 | | 0.545 | 0.645 | 0.529 | 0.57 |
| Tunisia | Low | 0.76 | 0.744 | | 0.736 | 0.769 | 0.529 | 0.727 |
| AUC | wellbeing | Sgb | mb | Mb int | glm | rf | nn | gbm |
| Chile | High | 0.655 | 0.717 | | 0.687 | 0.757 | 0.603 | 0.727 |
| Chile | Low | 0.830 | 0.802 | | 0.676 | 0.837 | 0.642 | 0.773 |
| Chile & Tunisia | High | 0.733 | 0.723 | 0.731 | 0.627 | 0.796 | 0.619 | 0.758 |
| Chile & Tunisia | Low | 0.763 | 0.733 | | 0.637 | 0.746 | 0.721 | 0.735 |
| Tunisia | High | 0.663 | 0.661 | | 0.537 | 0.710 | 0.616 | 0.658 |
| Tunisia | Low | 0.596 | 0.562 | | 0.579 | 0.614 | 0.492 | 0.579 |

**Table 1.** Predictive power for farm financial high and low wellbeing. Accuracy and Area Under the Curve (AUC) of all fitted models was evaluated on the test data from Chile and Tunisia. For corresponding receiver operator curves (see Extended Data Figure 4). For abbreviation explanation, see text.

### Climate change and farm financial wellbeing

First, we evaluated whether experiencing climate change had any impact on farm financial wellbeing. Decreasing rainfall and increasing temperatures were associated with reduced farm financial wellbeing (Figure 1). Combined, farmers in Chile and Tunisia, who have experienced reduced rainfalls, were significantly less likely to do financially well than farmers who did not experience reduced rainfalls (0.635, p=0.020). To a lesser extent, increases in temperature in the two countries also resulted in the likelihood farms to do financially well (0.751, p=0.11). Increasing drought frequencies and extreme weather experiences had no significant impact on farm financial wellbeing in any of the regions studied. The effects of increasing temperatures or decreasing rainfall were more discernible in Chile than Tunisia. Thus negative experiences with climate change lowered in some cases farm financial wellbeing, with the provision that the effects of the negative experiences may be country-specific.

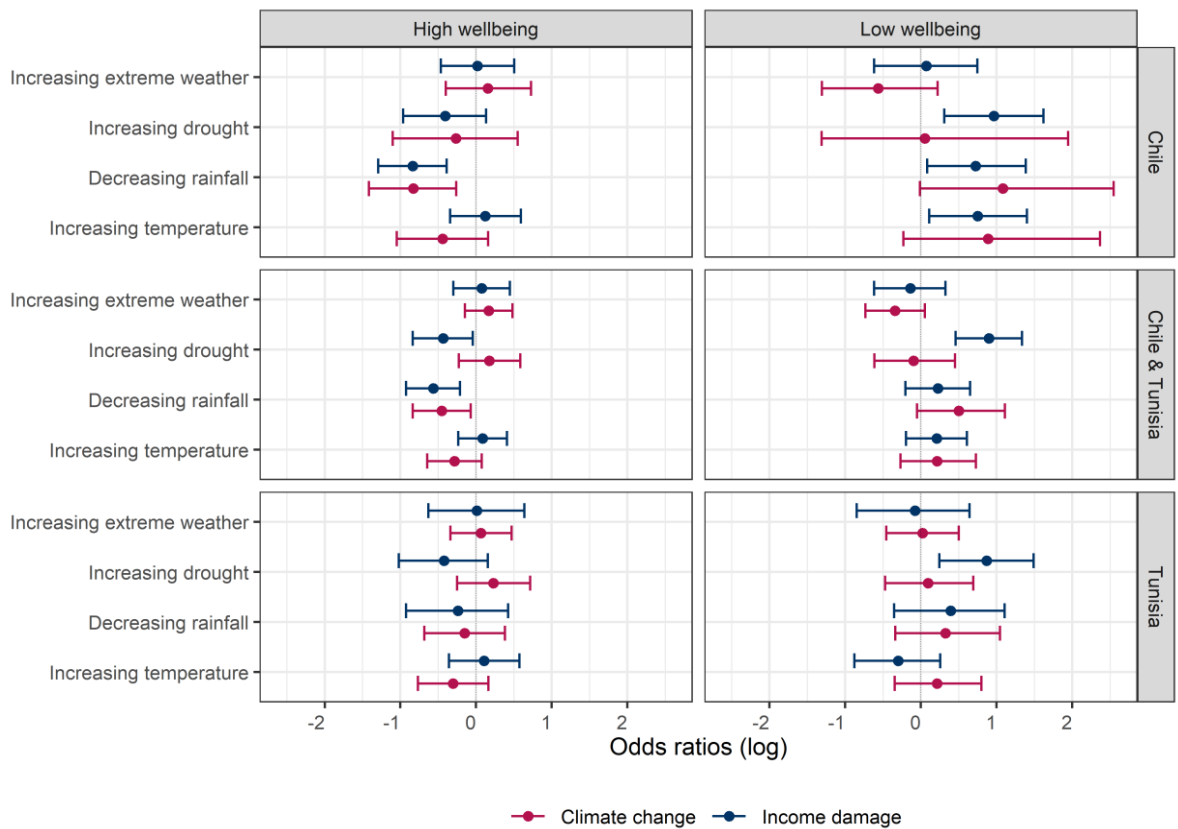

**Figure 1.** Effect of experiences with climate change and crop financial damage on financial wellbeing of a farm in Chile and Tunisia. Confidence intervals of the Odds-ratios (OR), based on logistic regression (see Extended Data Table 5 for more data).

Second, we investigated to what extent financial damage to crops, caused by specific climate change impacts, is associated with overall financial farm wellbeing. The results indicate that farms that performed financially well, the odds were that only decreasing rainfall associated income impacts were significantly associated with farm high wellbeing (0.568, p=0.002 for Chile and Tunisia combined, 0.434 , p<0.001 for Chile). Farms that were not doing financially well, the odds were that higher temperature associated income impacts were significantly associated with farms low wellbeing (2.119, p=0.021 for Chile) and more frequent drought (2.457, p<0.001 for Chile and Tunisia combined, 2.623, p=0.003 for Chile and 2.385, p=0.006 for Tunisia). Decreasing rainfall, especially in Chile, seemed to be somewhat relevant for explaining low wellbeing farms. It is noteworthy that although experiencing drought was not significantly associated with low or high financial wellbeing, the financial impacts of drought tended to be significantly associated with farm financial wellbeing.

### Variable importance for farm financial wellbeing

The sparse group boosting (sgb) algorism allowed the model to choose between individual and grouped independent variables for the predictive modeling (Figure 2). Arrow directions indicate the added effect size (log odds) of all variables within one group on the farm financial wellbeing, resulting in a latent variable. For high financial wellbeing, upward pointing arrows indicate that an overall increase of group variable values lead to an increased probability for high financial wellbeing while downward pointing arrows indicate a decreased probability of high wellbeing. Similarly, for low financial wellbeing, an arrow pointing upward means that increases in the group variable values increase the probability of low wellbeing. Thus higher/increasing social assets will increase the probability of farm high wellbeing. Note that no arrows were added for non-ordinal variables or groups of variables.

The most important predictors of farm high financial wellbeing, common both to Chile and Tunisia, are social (reliance on/use of information, trust in information sources, community, science or religion) and biophysical (farm size, water management systems used on the farm, diversity of crops used) assets as well as one individual variable, years of owning the farm (Figure 2). Natural assets (regional differences) are important predictors almost exclusively only for Chile. Prior farm ownership and the human asset group (including education, age, gender, knowledge) are important factors specific for Tunisia only. The most important predictors of farm low financial wellbeing, common both to Chile and Tunisia, are regional differences, income impact and financial asset groups, where for example increasing farm debt and reliance on orchard income increase the likelihood of farm low wellbeing. A number of factors increase the likelihood of farm low wellbeing in Tunisia only: these are social and biophysical assets groups, length of farm ownership, drought and varieties grown. For Chile only, the important individual factors are use of a well and years of farm management.

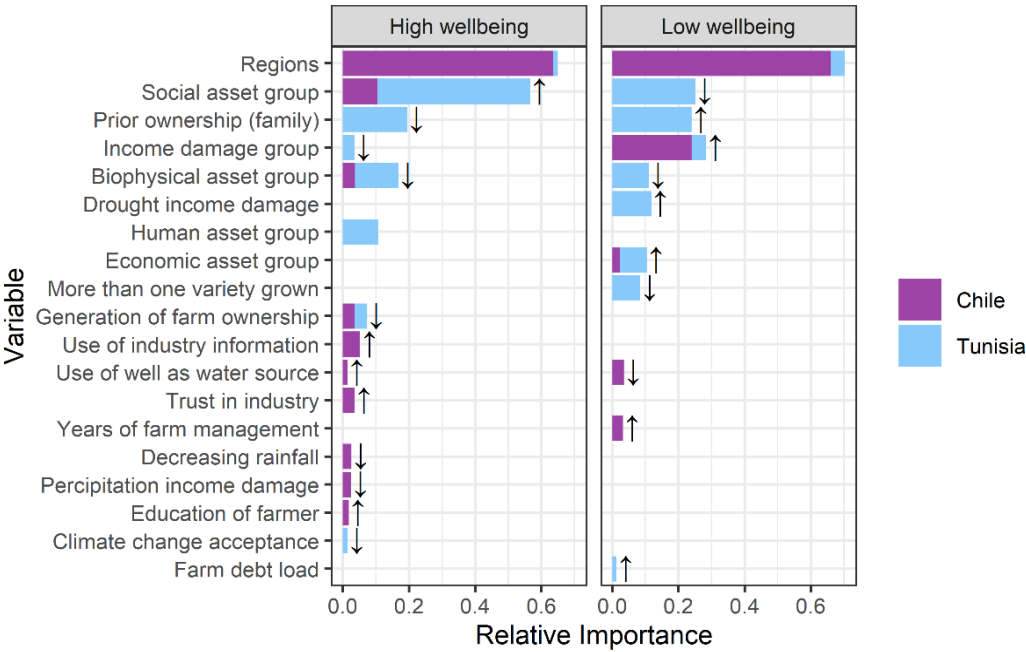

**Figure 2.** Most important variables contributing to farm financial wellbeing. Sparse group boosting model for Chile and Tunisia and high and low financially performing farms separately.

Note that some factors are important predictors of both high and low financial wellbeing, just with oposing effect. For example, increased well usage in Chile increases likelihood of high wellbeing while decreasing likelihood of low wellbeing. In Tunisia, prior family ownership decreases likelihood of high wellbeing while increasing the likelihood of low wellbeing. The exception are biophysical assets, that decrease the odds for high wellbeing and also decrease the odds for low wellbeing, indicating using biophysical assets, like adaptive measures, are only useful to help farmers with low financial wellbeing.

### Importance of variable interactions

We have examined whether interactions between independent variables may change the model outcomes vis a vie financial wellbeing of a farm (Figure 3). Even though the model that included variable interactions was not as predictive as the model including only additive effects (Table 1), the importance of each interaction still showcases interesting and important inter-dependencies in the datasets. One outcome is that the region variable seems to be less important when other interactions are considered. Interactions within and between social and human assets seem to be relevant for the farm's financial wellbeing, especially those related to use of information and trust. Interactions that involve adaptive measures, current assessment of climate change as well as education are also of relative importance. Such interactions point to inter-dependencies between variables and to likely confounding and mediating effects of certain variables.

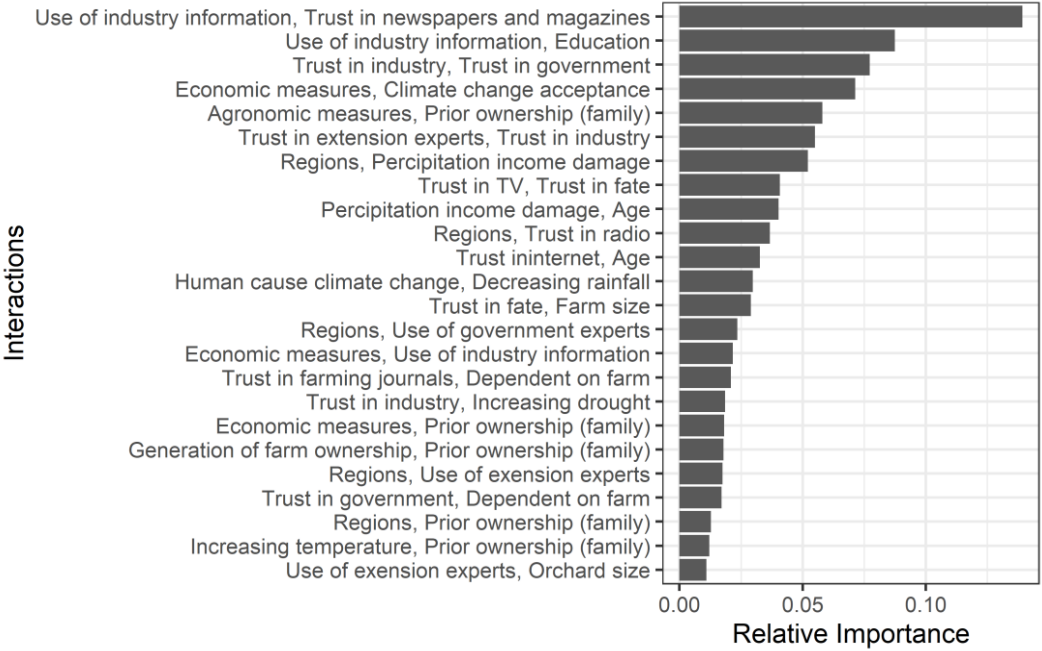

**Figure 3.** The most important interacting variables for farm financial wellbeing. Component-wise boosting model for Chile and Tunisia combined.

Figure 4 provides information about some noteworthy interactions that can affect farm financial wellbeing. Without the use of newspapers as a source on information the probability of high wellbeing drops in Chile and Tunisia when temperature increases or precipitation decreases (Figure 4, top left). However, when farmers used newspapers, financial wellbeing in Chile and Tunisia is not markedly reduced by increasing temperatures or decreasing precipitation (Figure 4, bottom left). Indeed, use of newspapers increased the probability of farm financial wellbeing irrespective whether or not temperature increases or precipitation decreases: the use of newspapers eliminated any negative effect of reduction in precipitation or increases in temperature on doing financially well. A similar effect was observed for trust in industry (Figure 4, top right and bottom right). Farmers, especially in Tunisia, who trusted industry as a source of information, were more likely to do financially well than farmers who did not trust industry, regardless whether or not they experienced reduction of precipitation. However, the effect of increasing temperatures on high wellbeing seems to be unchanged by trust in industry in Tunisia while in Chile trust in industry, compared to no trust in industry, intensified the negative effect of temperature increases on financial farm wellbeing.

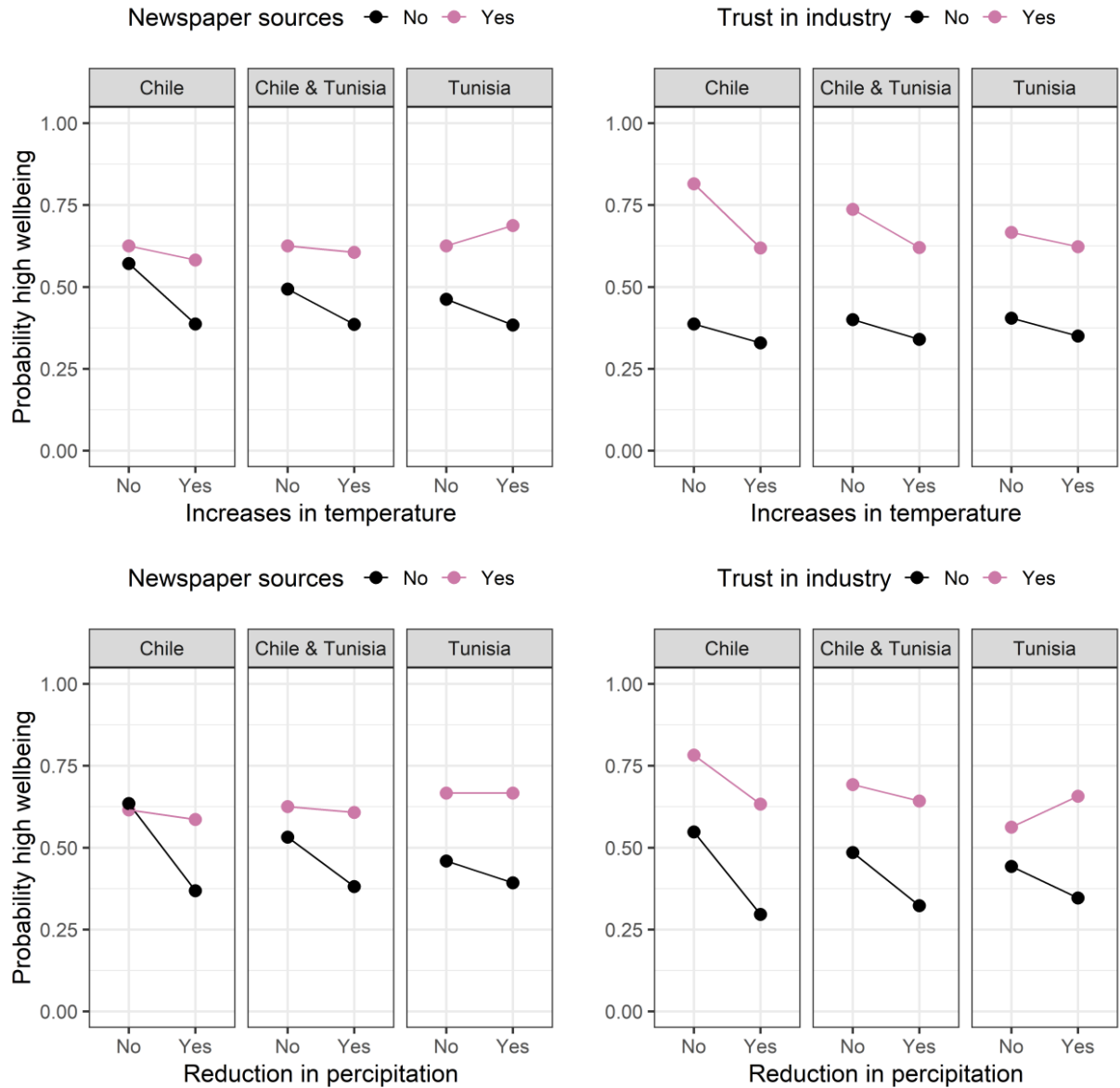

**Figure 4.** Probability for high financial wellbeing of the farm, based on an interaction between country, climate change factors, use of newspapers and trust in industry.

Trust in media, use of industry information and farm financial wellbeing indicate that farmers, regardless of their country of origin, who did not trust media and did not use information from industry had the lowest probability of doing financially well (Figure 5, top left). Farmers who did trust media sources but still did not use industry information, performed financially substantially better. Farmers with the highest probability of doing financially well were those that trusted the media and used industry information, where the trust factor acted synergistically with the use of information. The importance of trust for financial wellbeing can be illustrated with the effect of trust in industry, experts and government. Thus, trust in industry acted synergistically with trust in experts (Figure 5, top right) as did trust in government and trust in industry (Figure 5, bottom left).

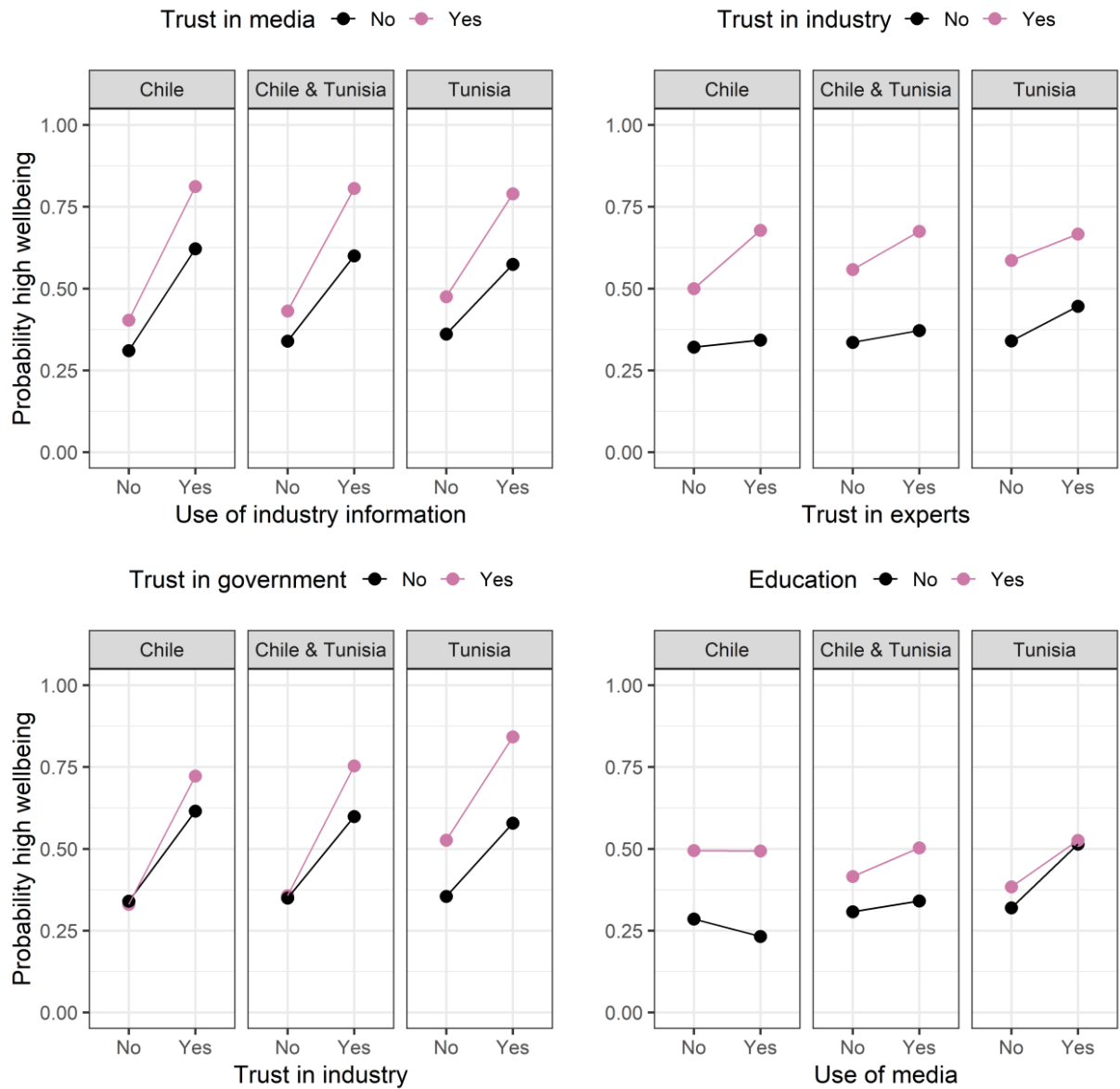

**Figure 5.** Probability for high financial wellbeing of the farm, based on an interaction between country, trust, use of information sources and education.

In all cases, farmers that trusted industry, experts or government were more likely to be financial well off than farmers who had no trust in their information sources. Other interactions, for example education and use of media also have a positive modifying effect on farm financial wellbeing in Chile but not in Tunisia: educated farmers who used media tended to be more likely to do well financially than farmers with low education (Figure 5, bottom right).

## Discussion

### Climate change and farm financial wellbeing

Previously published data indicate that climate change can have a detrimental effect on crop yields and farm income (3, 47, 58, 59, 60, 61, 62). Our findings suggest that only certain kinds of climate change can have an adverse effect on fruit farm financial wellbeing. Whereas odds are that increasing temperatures and reduction in the amount of rain will negatively affect fruit farm financial wellbeing, extreme climatic events do not seem to play such a role. Thus, while farmers have discussed possible fruit damage due to frost or hail events (63), such (still relatively rare) severe events do not appear to affect the mid to long-term farm income prospects. Indeed, fruit farmers are more likely to be concerned about drought issues (and consequently future water availability) (63), reflecting findings herein showing that increasing frequency of droughts had a negative effect on farm income and farm financial wellbeing.

However, based on the self-reported statements by farmers, at present climate change plays only a minor role in affecting fruit farm financial wellbeing. Indeed, variables other than climate change factors are better predictors of fruit farm financial wellbeing. In Chile, the location of a farm was a strong indicator of whether the farm is doing well financially, with farms in central Chile doing better than farms in Southern Chile. In Tunisia, farms more than one generation in the family possession, did worse financially. Chile and Tunisia also shared a number of important predictors. For example, for both Chile and Tunisia, access to information and trust of information sources are more important than climate change in predicting farm financial wellbeing: These shared factors are useful to predict both financially high and low performing farms: better the information access and more trust there is in information sources, better the farm financial performance and vice versa. On the other hand, we also observed that certain factors were more important either for farms doing financially poorly or doing financially well in Tunisia and Chile. Indeed, climate change factors play a more important role for farms not doing financially well. This indicates that farm specific circumstances need to be considered in order to improve or maintain farm financial performance. Consequently, different predictive factors may need to be used to address farms doing financially poorly or doing financially well. Finally, as eluded to above, not only individual factors but also synergistic effects of factors need to be considered as interactions between factors can affect their individual or combined importance for farm financial wellbeing. This is in agreement with other results, where financially healthier farms are more likely to be resilient against climatic impacts (48,64) and that inter-dependencies between factors can motivate responses to climate change (65).

Although the results presented herein indicate that climate change currently does not substantially impact fruit farm financial wellbeing, the situation may change in the future. This is evident from climate change trends analysed in this paper: the odds are that with higher temperatures and less precipitation fruit farm financial performance may decrease. Temperature predictions indicate continuing increases of winter night temperatures in Tunisia and Chile in the future. This will lead to winter chill deficits and potential problems with fruit tree phenology likely require changes in the type of fruit trees grown (14,66). Similarly, reduced precipitation and water availability in Tunisia has serious implications for the future. Much of the irrigation water for fruit trees comes from underground aqua ducts. If they are depleted or become saline, farmers in Tunisia may face major crop yield loses (67). In Chile, reduced water flow from the Andeas, increasing urbanization and inappropriate crop use could create water distribution bottlenecks (68,69). These latter predictions are in agreement with farmer climate predictions for the future: most worry about effects of drought and water availability (63). Resolution of these problems will require the implementation of specific adaptive measures that will reduce future vulnerability of fruit farms to climate change, measures that farmers and governments need to be willing to pay (64,70). In this respect, the regional specificity of some factors implicated in fruit farm financial wellbeing advocates for collecting extensive regional rather than country-wide datasets that rely on big data analysis.

**Use of hybrid modeling for climate change impacts on agriculture**

Predictive modeling provides new insights into data relationships that can serve to generate and test new hypotheses. We found that statistical models, utilizing limited datasets that reflect the requirements of relevant theories, can be used to make adequate predictions about the relationships between climate change, intervening variables, and the outcome variable. However, we have also shown that by combining statistical models with specific machine learning methods, such as boosting and cross-validation, we were able to substantially improve the predictability of the (generalized linear) statistical model. This hybrid model can still be interpreted through variable importance and odds ratios, but classical inference based on F and t statistics is not valid for variables selected through a data-driven process (71). Nevertheless, as already indicated, the hybrid model results do generate new insights that can be used to construct hypotheses to be tested by classical statistical means. Even though the random forest (a typical black box model) outperformed the sparse group boosting model, we believe that this improvement generally does not compensate for the loss of interpretability. With a similar analysis methodology, neural networks marginally outperformed logistic regression (72). The predictive sparse group boosting and component-wise boosting models were ultimately chosen for the current data analysis. The former model provided evidence of regional or supra-regional variables that are important for predicting whether fruit farms will perform financially well. The latter model revealed that the interaction between various variables and farm financial wellbeing, as the outcome variable, is not a simple one-to-one relationship. Rather, certain variables appear to have a modulating effect on variables that may directly affect the outcome variable.

Our experience with the hybrid model indicates that, especially when it is necessary to balance predictive improvements (usually requiring larger datasets) with loss of model interpretability, the research questions and the modeling tools available will dictate the extent and complexity of the data to be collected, whether the focus should be on regional or supra-regional datasets and the type and depth of analysis that can be performed. Machine learning provided the opportunity to include a broader range of independent variables with substantially better predictability of farm`s financial wellbeing and clarity of data presentation than offered by traditional regression analysis. We believe that, through group-component-wise boosting of generalized linear models, our hybrid approach can generate useful predictions in high dimensional settings, while still preserving basic interpretability, like variable importance and odds ratios. This way, new hypotheses and models can be generated, left to be validated or rejected by future research. The key challenge for future studies will be to find the correct balance between a theory-based approach, where a limited number of likely relevant variables are included in the survey design and resulting datasets, and a black-box approach that relies on deep mining of the largest possible number of data points.

## Data and code availability

The data and the corresponding code for conducting the analysis can be found in the supplement or on github (https://github.com/FabianObster/pasit_financial_wellbeing)

# References

1. IPCC, 2019. Summary for Policymakers. *In: Climate Change and Land: an IPCC special report on climate change, desertification, land degradation, sustainable land management, food security, and greenhouse gas fluxes in terrestrial ecosystems* [P.R. Shukla, J. Skea, E. Calvo Buendia, V. Masson-Delmotte, H.- O. Pörtner, D. C. Roberts, P. Zhai, R. Slade, S. Connors, R. van Diemen, M. Ferrat, E. Haughey, S. Luz, S. Neogi, M. Pathak, J. Petzold, J. Portugal Pereira, P. Vyas, E. Huntley, K. Kissick, M. Belkacemi, J. Malley, (eds.)]. Cambridge University Press, Cambridge, (2019).

2. IPCC, 2022. *Climate Change 2022: Impacts, Adaptation, and Vulnerability.* Contribution of Working Group II to the Sixth Assessment Report of the Intergovernmental Panel on Climate Change [Pörtner, H.-O; Roberts, D.C; Tignor, M; Poloczanska, E.S; Mintenbeck, K; Alegría, A; Craig, M;  Langsdorf, S; Löschke, S; Möller, V; Okem, A; Rama B. (eds.)]. Cambridge University Press, (2022).

3.  Zhao,C. *et al.* Temperature increase reduces global yields of major crops in four independent estimates. *PNAS* **114**, 9326-9331 (2017).

4. Fanzo, J.,  McLaren, R., Davis, C. & Choufani, J. How to ensure nutrition for everyone under climate change and variability.  GCAN policy notes 1. International Food Policy Research Institute (IFPRI) (2017).

5. Quero-García, J., Iezzoni, A., Pulawska,J., Lang, G.A. (Eds.). Cherries: Botany, Production and Uses. CABI. (2017).

6. Manganaris, G. A., Minas, I., Cirilli, M., Torres, R., Bassi, D., & Costa, G. Peach for the future: A specialty crop revisited. *Scientia Horticulturae* **305,** 111390 (2022).

7. Predieri, S., Dris, R., Sekse, L., & Rapparini, F. Influence of environmental factors and orchard management on yield and quality of sweet cherry. *J. Food Agric. Environ.* **1**, 263-266 (2003).

8. Jackson, D.I.  Climate and Fruit Plants. In *Temperate and subtropical Fruit production.* Jackson,D.I.,  Looney, N.F.,  Morley-Bunker, M. Ed. 3rd Edition. CAB International. pp. 11-17 (2011).

9. Ghrab, M., BenMimoun, M.,  Masmoudi, M.M. &  BenMechlia, N. The behaviour of peach cultivars under warm climatic conditions in the Mediterranean area. *Int J Environ Stud* **7,** 3–14 (2014).

10. Measham, P. F., Quentin, A. G., & MacNair, N. Climate, winter chill, and decision-making in sweet cherry production. *HortScience* **49,** 254-259 (2014).

11. Zhang, L., Ferguson, L., & Whiting, M. D.  Temperature effects on pistil viability and fruit set in sweet cherry. *Scientia Horticulturae* **241**, 8-17 (2018).

12. Sønsteby, A., & Heide,O.M. Temperature effects on growth and floral initiation in sweet cherry (Prunus avium L.). *Scientia Horticulturae* **257**, 108762, (2019).

13. Penso G.A., Citadin, I., Scariotto, S., Santos, C., Junior, A.W., Bruckner, C.H. & Rodrigo J. Development of Peach Flower Buds under Low Winter Chilling Conditions. *Agronomy* **10,** 428-448 (2020).

14. Fernandez, E., Whitney, C., Cuneo,I.F. & Luedeling,E. 2020. Prospects of decreasing winter chill for deciduous fruit production in Chile throughout the 21st century. *Climatic Change* **159**, 423-439 (2020).

15. Lopez, G. & DeJong, T.M. Spring temperatures have a major effect on early peach fruit growth. *J Hort Sci Biotech.* **82**, 507-512 (2007).

16. Usenik, V., & Stampar, F. The effect of environmental temperature on sweet cherry phenology. *European Journal of Horticultural Science*, **76**, 1-5 (2011).

17. Syvertsen, J. P. Integration of water stress in fruit trees. *HortScience* **20**, 1039-1043 (1985).

18. Alae-Carew, C., Nicoleau, S., Bird, F.A., Hawkins, P., Tuomisto, H.L., Haines, A., Dangour, A.D. & Scheelbeek, P.F.D. The impact of environmental changes on the yield and nutritional quality of fruits, nuts and seeds: a systematic review. *Environ Res Lett.* **15**, 023002 (2020).

19. Botzen, W.J.W., Bouwer, L.M. & van den Bergh, J.C.J.M. Climate change and hailstorm damage: Empirical evidence and implications for agriculture and insurance. *Res. En. Econ.* **32**, 341–362 (2010).

20. Seneviratne, S.I. *et al.* 2012: Changes in climate extremes and their impacts on the natural physical environment. In: Managing the Risks of Extreme Events and Disasters to Advance Climate Change Adaptation [Field, C.B. *et al* (eds.)]. A Special Report of Working Groups I and II of the Intergovernmenta I Panel on Climate Change (IPCC). Cambridge University Press, Cambridge, UK, and New York, NY, USA, pp. 109-230.

21. Lauren, E., Parker, A., McElrone, J., Ostoja,S.M & Forrestel, E.J., Extreme heat effects on perennial crops and strategies for sustaining future production. *Plant Science* **295**,110397 (2020).

22. *Nelson, G. et al.* Agriculture and climate change in global scenarios: why don't the models agree. *Agricultural Economics* **45,** 85-101 (2014).

23. Lampe, M. *et al.* Why do global long-term scenarios for agriculture differ? An overview of the AgMIP Global Economic Model Inter-comparison. *Agricultural Economics* **45**, 3-20 (2014).

24. R. Mendelsohn, R., Nordhaus, W.D. & Shaw, D. The impact of global warming on agriculture: a Ricardian analysis *Am. Econ. Rev.* **84**, 753-771 (1994).

25. Mendelsohn, R.O. & Massetti, E. The use of cross-sectional analysis to measure climate impacts on agriculture: Theory and evidence. *Review of Environmental Economics and Policy* **11**, 280-298 (2017).

26. Carter, C., Cui, X., Ghanem, D., & Mérel, P. Identifying the economic impacts of climate change on agriculture. *Annual Review of Resource Economics 10*, 361-380 (2018).

27. Kononenko, I. Machine learning for medical diagnosis: history, state of the art and perspective. *Artificial Intelligence in medicine* **23**, 89-109 (2001).

28. Bishop, C.M. Pattern Recognition and Machine Learning. Springer (2006).

29. Bzdok, D., Altman, A. & Krzywinski, M. Statistics versus Machine Learning. *Nat Methods* **15**, 233-234 (2018).

30. McQueen, R.J., Garner, S.R., Nevill-Manning, C.G. & Witten, I.H. Applying machine learning to agricultural data. *Comput. Electron. Agric.* **12**, 275–293 (1995).

31. González-Recio,O., Guilherme J.M., Rosa, G.J.M. & Gianola,D. Machine learning methods and predictive ability metrics for genome-wide prediction of complex traits. *Livestock Science* **166**, 217-231 (2014).

32. Coble, K. H., Mishra, A.K., Ferrell, S. & Griffin, T. Big Data in Agriculture: A Challenge for the Future. *Applied Economic Perspectives and Policy* **40**, 79-96 (2018).

33. Kamilaris, A. & Prenafeta-Boldú, F. X. Deep learning in agriculture: A survey. *Computers and electronics in agriculture* **147**, 70-90 (2018).

34. Liakos, K. G., Busato, P., Moshou, D., Pearson, S. & Bochtis, D. Machine learning in agriculture: A review. *Sensors* **18**, 2674 (2018).

35. Zhu, N.,*et al.* Deep learning for smart agriculture: Concepts, tools, applications, and opportunities. *International Journal of Agricultural and Biological Engineering* **11**, 32-44 (2018).

36. Ansarifar, J., Wang, L. & Archontoulis, S.V. An interaction regression model for crop yield prediction. *Sci Rep* **11**, 17754 (2021).

37. Tong, H. & Nikoloski, Z. Machine learning approaches for crop improvement: Leveraging phenotypic and genotypic big data. *Journal of Plant Physiology* **257**, 153354 (2021).

38. Jakariya, M. *et al.* Assessing climate-induced agricultural vulnerable coastal communities of Bangladesh using machine learning techniques. *Science of the Total Environment* **742**, 140255 (2020).

39. Avand, M. & Moradi, H. Using machine learning models, remote sensing, and GIS to investigate the effects of changing climates and land uses on flood probability. *Journal of Hydrology* **595**, 125663 (2021).

40. Guo, Y. *et al.* Integrated phenology and climate in rice yields prediction using machine learning methods. *Ecological Indicators* **120**, 106935 (2021).

41. Rolnick,D. *et al.* Tackling climate change with Machine Learning. *ACM Computing Surveys* **55**, Article 42 (2022).

42. IPCC. Climate Change 2022: Impacts, Adaptation, and Vulnerability. Contribution of Working Group II to the Sixth Assessment Report of the Intergovernmental Panel on Climate Change [Pörtner, H.O. et al (eds.)]. Cambridge University Press. In Press (2022).

43. Crane-Droesch, A. Machine learning methods for crop yield prediction and climate change impact assessment in agriculture. *Environmental Research Letters* **13**, 114003 (2018).

44. Van Klompenburg, T., Kassahun, A. & Catal, C. Crop yield prediction using machine learning: A systematic literature review. *Computers and Electronics in Agriculture* **177**, 105709 (2020).

45. Meshram, V. *et al.* Machine learning in the agricultural domain: state-of-art survey. *Artificial Intelligence in the Life Sciences* **1**, 100010 (2021).

46. Mark, H.S. *et al.* Adapting agriculture to climate change. *PNAS* **104**, 19691-19696 (2007).

47. Khan, T., Sherazi, H.H.R., Ali, M., Letchmunan, S. & Butt, U.M. Deep Learning-Based Growth Prediction System: A Use Case of China Agriculture. *Agronomy* **11***,* 1551 (2021).

48.  Shahhosseini, M., Hu, G., Huber, I. & Archontoulis, S.V. Coupling machine learning and crop modeling improves crop yield prediction in the US Corn Belt. *Sci. Rep.* **11**, 1606 (2021).

49. Porritt, J. The World in Context: beyond the business case for sustainable development. University of Cambridge Programme for Industry. (2003). www.cisl.cam.ac.uk/publications/the-world-in-context. Accessed 12/09/14.

50. Ivory, S. & Brooks, S.B. An updated conceptualisation of corporate sustainability: Five resources sustainability. In Proceedings. British Academy of Management (BAM), British Academy of Management Annual Conference 2018, Bristol, United Kingdom (2018).

51. Hothorn, T., Buehlmann, P., Kneib, T., Schmid, M. & Hofner, B.  Mboost: Model-Based Boosting. R package version 2.9-7 (2022). https://CRAN.R-project.org/package=mboost.

52. Tibshirani, R. Regression Shrinkage and Selection Via the Lasso. *Journal of the Royal Statistical Society: Series B (Methodological)* **58**, 267-288 (1996).

53. Obster, F. & Heumann, C.  Sparse-group boosting - Unbiased group and variable selection. Preprint at https://arxiv.org/abs/2206.06344 (2022).

54. Nelder, J. A. R. & Wedderburn, W. M. Generalized Linear Models.  *Journal of the Royal Statistical Society. Series A (General)* **135**, 370–84 (1972).

55. Hofner, B. *et al.* Model-based boosting in R: a hands-on tutorial using the R package mboost. *Comput Stat* **29**, 3–35 (2014).

56. Simon, N., Friedman, J., Hastie, T. & Tibshirani, R.  A Sparse-Group Lasso. *Journal of Computational and Graphical Statistics* **22**, 231-245 (2013).

57. Wickham, H.  ggplot2: Elegant Graphics for Data Analysis. Springer-Verlag New York. (2016).  https://ggplot2.tidyverse.org.

58. Lobell,D.B. &  Gourdji, S.M. The Influence of Climate Change on Global Crop Productivity. *Plant Physiology* **160**, 1686 -1697 (2012).

59. Bobojonov, I.  & Aw-Hassan, A. Impacts of climate change on farm income security in Central Asia: An integrated modeling approach. *Agriculture, Ecosystems & Environment* **188,** 245-255 (2014).

60. Abraham, T.W. & Fonta, W.M. Climate change and financing adaptation by farmers in northern Nigeria. *Financ Innov* **4,** 11 (2018).

61. Dalhaus, T. *et al.* The Effects of Extreme Weather on Apple Quality. *Sci Rep* **10,** 7919 (2020).

62. El Yaacoubi, A. *et al.* Potential vulnerability of Moroccan apple orchard to climate change–induced phenological perturbations: effects on yields and fruit quality. *Int J Biometeorol* **64**, 377–387 (2020).

63. Pechan, P., Bohle, H. & Obster, F.  Reducing vulnerability of orchards to climate change impacts. (2023).

64. Pechan, P.  Bohle, H.  & Obster, F. Factors affecting cherry and peach farming community response  to climate change. (2023).

65. van Valkengoed, A.M.  & Steg, L. Meta-analyses of factors motivating climate change adaptation behaviour. *Nature Clim Change* **9**, 158–163 (2019).

66. Benmoussa, L., Luedeling, E., Ghrab, M., & Ben Mimoun, M. Severe winter chill decline impacts Tunisian fruit and nut orchards *Climatic Change* **162**, 1249–1267 (2020).

67. Verner, D., Treguer, D., Redwood, J., Christensen, J., McDonnell, R., Elbert, C. & Konishi, Y. Climate Variability, Drought, and Drought Management in Tunisia's Agricultural Sector. World bank Group, 114 pp, (2018).

68. Meza,F..J., Wilks, D.S., Gurovich, L. & Bambach, N. Impacts of Climate Change on Irrigated Agriculture in the Maipo Basin, Chile: Reliability of Water Rights and Changes in the Demand for Irrigation. *J. Water Resour. Plann. Manage.* **138,** 421- 430 (2012).

69. Novoa, V., Ahumada-Rudolph,R., Rojas, O., Sáez, K., de la Barrera, F. &  Arumí, J.L. Understanding agricultural water footprint variability to improve water management in Chile. *Science of The Total Environment* **670,** 188-199 (2019).

70. OECD. Water and Climate Change Adaptation: Policies to Navigate Uncharted Waters. *Studies on Water*, OECD Publishing  (2013).

71. Berk, R., Brown, L., Buja, A., Zhang, K. & Zhao, L. Valid Post-Selection Inference. *The Annals of Statistics* **41**, 802–837 (2013).

72. Raval, M. et al. Automated predictive analytic tool for rainfall forecasting. *Sci Rep* **11**, 17704 (2021).

## Acknowledgements

The authors would like to thank members of the PASIT project for their feedback during survey preparation.

## Author contributions

## Funding

This research was conducted within the project "Phenological And Social Impacts of Temperature Increase – climatic consequences for fruit production in Tunisia, Chile and Germany" (PASIT; grant number 031B0467B of the German Federal Ministry of Education and Research). Open Access funding was enabled by LMU.

## Competing interests

The authors declare no competing interests.

# Extended Data Figures

## 1. Overview of all independent variables and corresponding groups used in the analysis     *numbers in bracket indicate the reference category

| Variable name | Category* | n | Group | Original Questionnaire scale |
|---|---|---|---|---|
| Agronomic measures | yes (no) | 647 | biophysical asset | Dichotomous |
| Economic measures | yes (no) | 464 | biophysical asset | dichotomous |
| Technological measures | Yes (no) | 721 | biophysical asset | dichotomous |
| Use of well as water source | Yes (no) | 231 | biophysical asset | dichotomous |
| Farm size | >7ha (≤7ha) | 283 | biophysical asset | Interval |
| Orchard size | >2ha (≤2ha) | 318 | biophysical asset | Interval |
| More than one variety grown | Yes (no) | 508 | biophysical asset | dichotomous |
| Other products | Yes (no) | 571 | biophysical asset | dichotomous |
| Regions | CentralChile | 200 | natural asset | Nominal |
| Regions | CentralTunisia | 200 | natural asset | Nominal |
| Regions | NorthernTunisia | 201 | natural asset | Nominal |
| Regions | SouthernChile | 200 | natural asset | Nominal |
| Percentage of income invested | ≥80% (<80%) | 137 | economic asset | Dichotomous |
| Farm debt load | Heavy in debt (rest) | 96 | economic asset | Dichotomous |
| Family members dependent on farm | >2 (≤2) | 528 | economic asset | Count |
| Family farm engagement | >2 (≤2) | 203 | economic asset | Count |
| Climate change acceptance | Yes (no) | 676 | human asset | dichotomous |
| Human cause climate change | Yes (no) | 685 | human asset | dichotomous |
| Climate change causes extremes | Yes (no) | 755 | human asset | dichotomous |
| Age of farmer | >50 (≤50) | 438 | human asset | Count |
| Gender of farmer | M (F) | 680 | human asset | dichotomous |
| Education of farmer | ≥ primary (no primary) | 632 | human asset | dichotomous |
| Generations of farm ownership | ≥3 (0) | 218 | human asset | Count |
| Generations of farm ownership | 2 (0) | 130 | human asset | Count |
| Generations of farm ownership | 1 (0) | 229 | human asset | Count |
| Prior ownership | Family (other) | 399 | human asset | dichotomous |
| Years of farm management | >10 (≤10) | 437 | human asset | Count |
| Use of newspapers and magazines | 4,5 (1,2,3) | 95 | social asset | Likert 5 point |
| Use of farming journals | 4,5 (1,2,3) | 161 | social asset | Likert 5 point |
| Use of television | 4,5 (1,2,3) | 415 | social asset | Likert 5 point |
| Use of radio | 4,5 (1,2,3) | 219 | social asset | Likert 5 point |
| Use of internet | 4,5 (1,2,3) | 319 | social asset | Likert 5 point |
| Use of extension experts | 4,5 (1,2,3) | 346 | social asset | Likert 5 point |
| Use of government experts | 4,5 (1,2,3) | 166 | social asset | Likert 5 point |
| Use of neighbours | 4,5 (1,2,3) | 313 | social asset | Likert 5 point |
| Use of industry information | 4,5 (1,2,3) | 192 | social asset | Likert 5 point |
| Use of farm associations | 4,5 (1,2,3) | 97 | social asset | Likert 5 point |
| Trust in newspapers and magazines | 4,5 (1,2,3) | 174 | social asset | Likert 5 point |
| Trust in farming journals | 4,5 (1,2,3) | 291 | social asset | Likert 5 point |
| Trust in television | 4,5 (1,2,3) | 329 | social asset | Likert 5 point |
| Trust in radio | 4,5 (1,2,3) | 241 | social asset | Likert 5 point |
| Trust in internet | 4,5 (1,2,3) | 319 | social asset | Likert 5 point |
| Trust in extension experts | 4,5 (1,2,3) | 433 | social asset | Likert 5 point |
| Trust in government workers | 4,5 (1,2,3) | 268 | social asset | Likert 5 point |
| Trust in neighbours | 4,5 (1,2,3) | 319 | social asset | Likert 5 point |
| Trust in industry | 4,5 (1,2,3) | 215 | social asset | Likert 5 point |
| Trust in farm associations | 4,5 (1,2,3) | 184 | social asset | Likert 5 point |
| Trust in government institutions | 4,5 (1,2,3) | 213 | social asset | Likert 5 point |
| Trust in other farmers | 4,5 (1,2,3) | 168 | social asset | Likert 5 point |
| Trust in my religion | 4,5 (1,2,3) | 236 | social asset | Likert 5 point |
| Trust in fate | 4,5 (1,2,3) | 268 | social asset | Likert 5 point |
| Temperature increase | Yes (no) | 629 | climate experience | Dichotomous |
| Rainfall decrease | Yes (no) | 659 | climate experience | Dichotomous |
| Drought increase | Yes (no) | 671 | climate experience | Dichotomous |
| Extreme weather increase | Yes (no) | 542 | climate experience | Dichotomous |
| Temperature income damage | 4,5 (1,2,3) | 294 | income damage | Likert 5 point |
| Precipitation income damage | 4,5 (1,2,3) | 219 | income damage | Likert 5 point |
| Drought income damage | 4,5 (1,2,3) | 161 | income damage | Likert 5 point |
| Extreme weather income damage | 4,5 (1,2,3) | 187 | income damage | Likert 5 point |

*numbers in bracket indicate the reference category

**Extended Data Figure 1 | Overview of all independent variables and corresponding groups used in the analysis**

| Variable name | Question |
|---|---|
| Agronomic measures | From the following list, please select the adaptive measures you have already undertaken in the past 10 years to reduce the impact of climatic changes on your farming operations. More than one answer is possible. *a.Changed tree thinning and pruning practices to reduce hail and rain damage  b. Used chemical treatments for bud breaking and control flowering time c. Planted/re-grafted new varieties that have low chilling requirements d. Planted/re-grafted drought tolerant varieties e. Planted/re-grafted early or late maturing varieties* |
| Technologic measures | From the following list, please select the adaptive measures you have already undertaken in the past 10 years to reduce the impact of climatic changes on your farming operations. More than one answer is possible. *b. Used chemical treatments for bud breaking and control flowering time  f. Installed canopy (nets) against hail, sun and heat damage g. Installed water irrigation systems (for example drip irrigation) h. Improved the efficiency of irrigation systems to reduce water use and energy costs* |
| Economic measures | From the following list, please select the adaptive measures you have already undertaken in the past 10 years to reduce the impact of climatic changes on your farming operations. More than one answer is possible. *i. Took land out of production j. Sold or rented part or all of the farm property k. Purchased crop damage insurance l. Got an off-farm job to supplement farm income (you and/or your spouse)* |
| Use of well as water source | Which water sources are used on your farm for agriculture (for example well, river water) |
| Farm size | Farm size (ha) |
| Orchard size | Size of your peach/cherry orchard (ha) |
| More than one variety grown | Varieties grown |
| Other products | Other farm products |
| Regions | Central Chile |
| Regions | Southern Chile |
| Regions | Northern Tunisia |
| Regions | Central Tunisia |
| Percentage of income invested | What percentage of your annual farm brutto income have you invested into adaptive measures in recent years to reduce the impact of climate change on your farm operations? *5 point Likert scale from 0-20% to 80-100%* |
| Farm debt load | How much in debt is your farm business? No debt, lightly in debt, moderately in debt, heavily in debt |
| Family members dependent on farm | Number of family members dependent on the farm activities |
| Family farm engagement | Number of family members working on your farm |
| Climate change acceptance | Global climate is not changing *Correct, incorrect.* |
| Human cause climate change | Human activities, such as burning of fossil fuels, are an important cause of climate change *Correct, incorrect* |
| Climate change causes extremes | Climatic changes can lead to an increased intensity and frequency of extreme weather events, such as hail, floods, frost and high winds *Correct, incorrect* |
| Age of farmer | Age |
| Gender of farmer | Gender *Male, Female* |
| Education of farmer | Education *1.None, 2. Incomplete primary, 3. Primary, 4. Incomplete secondary, 5. Complete secondary, 6. Technical education, (specify subject area), 7. University (specify subject area), 8. University postgraduate (specify subject area), 9. Other* |
| Generations of farm ownership | How many generations does your family own this farm? |
| Prior ownership | Who owned the farm before you? |
| Years of farm management | How many years have you been managing this farm? |
| Use of newspapers and magazines | How often do you use the following information sources on how to deal with climate change impacts on agricultural production? National newspapers or magazines *5 point Likert scale from not at all to very often* |
| Use of farming journals | Farming journals *5 point Likert scale from not at all to very often* |
| Use of television | TV *5 point Likert scale from not at all to very often* |
| Use of radio | Radio *5 point Likert scale from not at all to very often* |
| Use of internet | Internet *5 point Likert scale from not at all to very often* |
| Use of extension experts | Farm extension workers *5 point Likert scale from not at all to very often* |
| Use of government experts | Government experts/experts *5 point Likert scale from not at all to very often* |
| Use of neighbours | Neighbours/communities *5 point Likert scale from not at all to very often* |
| Use of industry information | Agriculture industry/export industry *5 point Likert scale from not at all to very often* |
| Use of farm associations | Farmer associations/cooperatives *5 point Likert scale from not at all to very often* |
| Trust in newspapers and magazines | How much do you trust the following information sources to provide you with reliable information on how protect your farm against possible climate change impacts? National newspapers or magazines *5 point Likert scale from not at all to very often* |
| Trust in farming journals | Farming journals *5 point Likert scale from not at all to very often* |
| Trust in television | TV *5 point Likert scale from not at all to very often* |
| Trust in radio | Radio *5 point Likert scale from not at all to very often* |
| Trust in internet | Internet *5 point Likert scale from not at all to very often* |
| Trust in extension experts | Farm extension workers *5 point Likert scale from not at all to very often* |
| Trust in government workers | Government experts/experts *5 point Likert scale from not at all to very often* |
| Trust in neighbours | Neighbours/communities *5 point Likert scale from not at all to very often* |
| Trust in industry | Agriculture industry/export industry *5 point Likert scale from not at all to very often* |
| Trust in farm associations | Farmer associations/cooperatives *5 point Likert scale from not at all to very often* |
| Trust in government institutions | I trust government institutions to help me to protect my farm against future impacts of climate change *5 point Likert scale from strongly disagree to strongly agree* |
| Trust in other farmers | I trust other farmers to advise me on what adaptive measures I should select to reduce future impacts of climate change on my farm *5 point Likert scale from strongly disagree to strongly agree* |
| Trust in my religion | I trust my religion more than science to guide me how to protect my farm against future impacts of climate change |
| Trust in fate | I trust in fate to guide me how to protect my farm against future impacts of climate change  *5 point Likert scale from strongly disagree to strongly agree* |
| Temperature increase | In recent years, I have observed that the temperature on my farm  *1.has increased, 2. has not changed, 3. has decreased, 4. has become unpredictable* |
| Rainfall decrease | In recent years, I have observed that the rainfall on my farm  *1.has increased, 2. has not changed, 3. has decreased, 4. has become unpredictable* |
| Drought increase | In recent years, I have observed that the dry periods and drought on my farm  *1.has increased, 2. has not changed, 3. has decreased, 4. has become unpredictable* |
| Extreme weather increase | In recent years, I have observed that the extreme weather events on my farm *1.has increased, 2. has not changed, 3. has decreased, 4. has become unpredictable* |
| Temperature income damage | If you deal with damage to peaches/cherries on your farm in recent years due to changes in temperature during the fruit growing season, how serious would you rate the impact of the crop damage on your farm income that  year(s)? *5 point Likert scale from not at all serious to very serious* |
| Winter temperature income damage | If you deal with damage to peaches/cherries on your farm in recent years due to changes in temperature during the winter tree dormancy period, how serious would you rate the impact of the crop damage on your farm income that  year(s)?  *5 point Likert scale from not at all serious to very serious* |
| Precipitation income damage | If you deal with damage to peaches/cherries on your farm in recent years due to changes in rainfall during the fruit growing season, how serious would you rate the impact of the crop damage on your farm income that  year(s)? *5 point Likert scale from not at all serious to very serious* |
| Drought income damage | If you deal with damage to peaches/cherries on your farm in recent years due to changes in dry periods and droughts during the fruit growing season, how serious would you rate the impact of the crop damage on your farm income that  year(s)? *5 point Likert scale from not at all serious to very serious* |
| Extreme weather income damage | If you deal with damage to peaches/cherries on your farm in recent years due to changes in extreme weather events during the fruit growing season, how serious would you rate the impact of the crop damage on your farm income that  year(s)? *5 point Likert scale from not at all serious to very serious* |

**Extended Data Figure 2 | Relevant survey questions utilized for this manuscript**

| Model | Short name | package | Main hyperparameters | Interpretability |
|---|---|---|---|---|
| Logistic regression | glm | base | - | interpretable |
| Model-based boosting without interactions | mb | mboost | Nu=0.3, mstop = 3000 | interpretable |
| Sparse group boosting | sgb | mboost | Nu=0.3, alpha = 0.5, mstop = 3000 | Interpretable |
| Model-based boosting with interactions | mb int | mboost | Nu= 0.3, mstop = 3000 | Interpretable |
| Random forest | Rf | randomForest | Ntree=500, mtry=7 | Not interpretable |
| Gradient boosting machines | gbm | gbm | Trees=100, interaction.depth=3 | Not interpretable |
| Neural Network | nn | neuralnet | hidden layers: 1, logistic activation | Not interpretable |

**Extended Data Figure 3 | Overview of the models used in the data analysis**.

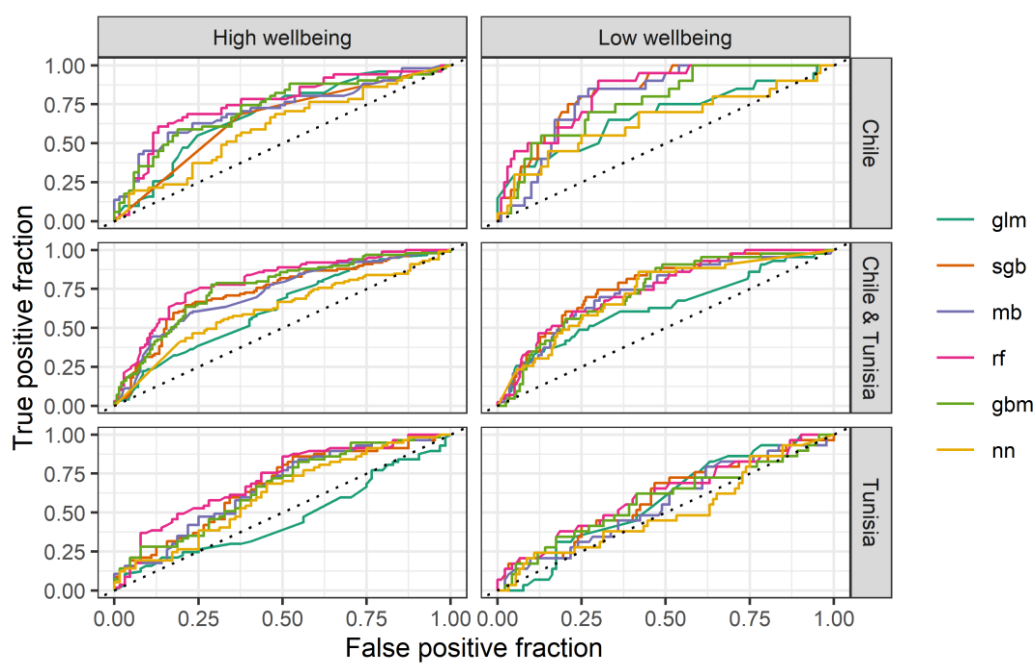

**Extended Data Figure 4 | Overview of ROC curves for selected models using the Chile and Tunisia datasets**

| | Chile & Tunisia | Chile & Tunisia | Chile | Chile | Tunisia | Tunisia |
|---|---|---|---|---|---|---|
| | Odds ratio (p-value) | Odds ratio (p-value) | Odds ratio (p-value) | Odds ratio (p-value) | Odds ratio (p-value) | Odds ratio (p-value) |
| **Climate Change experience** | High wellbeing | Low wellbeing | High wellbeing | Low wellbeing | High wellbeing | Low wellbeing |
| Increasing temperatures | 0.751 (0.11) | 1.238 (0.400 ) | 0.644 (0.150) | 2.428 (0.167) | 0.739 (0.201) | 1.237 (0.466) |
| Decreasing rainfall | 0.635 (0.020) | 1.650 (0.089) | 0.437 (0.004) | 2.950 (0.086) | 0.860 (0.577) | 1.384 (0.353) |
| Increasing drought frequency | 1.29 (0.394) | 0.906 (0.715) | 0.766 (0.523) | 1.052 (0.949) | 1.257 (0.353) | 1.099 (0.749) |
| Increasing extreme weather frequency | 1.18 (0.302) | 0.709 (0.086) | 1.170 (0.584) | 0.569 (0.146) | 1.068 (0.751) | 1.022 (0.930) |
| **Income impact** | | | | | | |
| Increasing temperatures | 1.092 (0.592) | 1.234 (0.305) | 1.131 (0.606) | 2.119 (0.021) | 1.113 (0.652) | 0.741 (0.299) |
| Decreasing rainfall | 0.568 (0.002) | 1.254 (0.297) | 0.434 (<0.001) | 2.064 (0.028) | 0.789 (0.489) | 1.481 (0.289) |
| Increasing drought frequency | 0.647 (0.031) | 2.457 (<0.003) | 0.664 (0.143) | 2.623 (0.003) | 0.656 (0.158) | 2.385 (0.006) |
| Increasing extreme weather frequncy | 1.08 (0.699) | 0.871 (0.565) | 1.019 (0.940) | 1.074 (0.837) | 1.009 (0.976) | 0.926 (0.838) |

**Extended Data Figure 5 | Financial wellbeing of a farm in Chile and Tunisia. Odds ratios and corresponding p-values based on logistic regression**